\documentclass{article}

\usepackage{graphicx}
\usepackage[hidelinks]{hyperref}

    \usepackage[final]{tccml_neurips_2020}

\usepackage[utf8]{inputenc} 
\usepackage[T1]{fontenc}    
\usepackage{hyperref}       
\usepackage{url}            
\usepackage{booktabs}       
\usepackage{amsfonts}       
\usepackage{nicefrac}       
\usepackage{microtype}      

\title{High-resolution global irrigation prediction with Sentinel-2 30m data}

\author{%
  Weixin Wu \\
  UC Berkeley \\
  \texttt{wwu34@berkeley.edu} \\
  \AND
  Sonal Thakkar \\
  UC Berkeley \\
  \texttt{sonalthakkar@berkeley.edu} \\
  \And
  Will Hawkins \\
  UC Berkeley \\
  \texttt{whawkins@berkeley.edu} \\
  \And
  Hossein Vahabi \\
  UC Berkeley \\
  \texttt{puyavahabi@berkeley.edu} \\
  \And 
  Alberto Todeschini \\
  UC Berkeley \\
  \texttt{todeschini@berkeley.edu}
}

\begin{document}

\maketitle

\begin{abstract}
An accurate and precise understanding of global irrigation usage is crucial for a variety of climate science efforts. Irrigation is highly energy-intensive, and as population growth continues at its current pace, increases in crop need and water usage will have an impact on climate change. Precise irrigation data can help with monitoring water usage and optimizing agricultural yield, particularly in developing countries. Irrigation data, in tandem with precipitation data, can be used to predict water budgets as well as climate and weather modeling. With our research, we produce an irrigation prediction model that combines unsupervised clustering of Normalized Difference Vegetation Index (NDVI) temporal signatures with a precipitation heuristic to label the months that irrigation peaks for each cropland cluster in a given year. We have developed a novel irrigation model and Python package ("Irrigation30") to generate 30m resolution irrigation predictions of cropland worldwide. With a small crowdsourced test set of cropland coordinates and irrigation labels, using a fraction of the resources used by the state-of-the-art NASA-funded GFSAD30 project with irrigation data limited to India and Australia, our model was able to achieve consistency scores in excess of 97\% and an accuracy of 92\% in a small geo-diverse randomly sampled test set.

\end{abstract}

\section{Introduction}

Understanding global irrigation patterns, both the extent of cropland and water usage, can enable various climate science efforts. Irrigated cropland accounts for 70\% of global freshwater use (R. Q. Grafton et al., 2018). As population growth continues, the cropland required to keep up with food demand will significantly increase greenhouse gas productions (Tillman et al., 2011). This cropland demand will be particularly located in developing countries, where population and consumption growth are disproportionate. Furthermore, precise irrigation data enables various climate science applications like crop water productivity modeling (Deryng et al., 2016) and global biomass estimation (Mauser et al., 2015). To sustainably meet this existential challenge, a precise and accurate understanding of global irrigation usage is required.

Despite the need for accurate global irrigation data, current data sources are outdated, low-resolution, and potentially inaccurate. With our research, we develop a machine learning model that estimates irrigation at 30m resolution using the latest satellite and climate datasets in years 2017+. We also deliver a Python package\footnote{Python Package can be found here: \texttt{https://github.com/AngelaWuGitHub/irrigation30}} that allows researchers to make use of our irrigation model and generate irrigation predictions for small regions at a time.

\section{Related Work}

Meier et al. (2018) created a global irrigation map at 30 arcsec (approx. 0.9km at the Equator) through downscaling a lower-resolution statistically-based global irrigation map and analyzing remote sensing data. However, the median crop field size is far less than 1km$^2$ in the mainland United States (0.278km$^2$), and even smaller in Asia and Africa (Yan et al., 2016; Fritz et al., 2015). To accurately measure the irrigation extent, a higher-resolution mapping is needed on a global scale. 

The Global Food Security-support Analysis Data @ 30-m (GFSAD30) project provides irrigation labels at 30m resolution, although these irrigation products are currently limited to a few regions and data is accessible by visualization only (i.e., no downloadable content).\footnote{\url{https://croplands.org/app/map}} An earlier effort from the GFSAD team generated five-class global cropland extent, which marked irrigation and rainfed extent at 1km resolution for nominal year 2010 (Thenkabail et al., 2016). Xie et al. (2019) used a semi-automatic training approach to map irrigated cropland extent across the conterminous United States at 30m resolution. At the time of writing, there are no publications that produce 30m resolution irrigation mappings on a global scale.

\section{Research Methodology}

\subsection{Data Sources}

While many recent publications on irrigation detection used Landsat satellite data (Yan et al., 2016; Xie et al., 2017; Oliphant et al., 2019; Gumma et al., 2020; Phalke et al., 2020), Sentinel-2 satellite data was chosen for this analysis because of its higher revisit frequency (a five-day revisit frequency over land surface as compared to 16-day revisit frequency for Landsat 8). The shorter revisit period of Sentinel-2 allows monthly time series data to be more representative and less influenced by factors that cause outliers like cloud coverage. Sentinel-2 multispectral instrument covers 13 bands at resolutions ranging from 10 to 60 meters. We use NDVI, the normalized difference between near-infrared band and red band, as a measure of cropland vegetation coverage and its healthiness. 

Climate data, such as precipitation, is also used in our analysis. TerraClimate provides global monthly climate data at 2.5 arc minutes, which is approx. 4.2km at the Equator (Abatzoglou et al., 2018). 

We use the GFSAD30 global cropland map at 30m resolution to remove non-cropland from our analysis (Xiong et al., 2017; Teluguntla et al., 2017; Oliphant et al., 2017; Gumma et al., 2017; Phalke et al., 2017; Massey et al., 2017; Zhong et al., 2017; Congalton et al., 2017). The global GFSAD30 cropland products define areas of cropland, non-cropland, and water bodies. 

We use Google Earth Engine (GEE) to manage and process all data and analysis for this research. Sentinel-2 and TerraClimate are accessible via GEE, while GFSAD30 cropland data was manually uploaded as an asset to GEE. 

Credible and publicly available irrigation-labelled data is hard to come by. Other research teams have collected independent validation data to complement the publicly available crowdsourced labelled data in their analyses (Congalton et al., 2017; Xie et al., 2019). We use crowdsourced irrigation data from croplands.org\footnote{\url{https://croplands.org/app/data/search}} for evaluation but not model training, because their data does not provide complete global coverage.

\subsection{Exploration}

We chose an unsupervised approach to identify irrigated cropland. Clustering analysis is a commonly used technique to identify groups of objects sharing similar patterns. Land use patterns can be observed through the change in NDVI over a 12-month period. 




\begin{figure}
  \includegraphics[width=\linewidth]{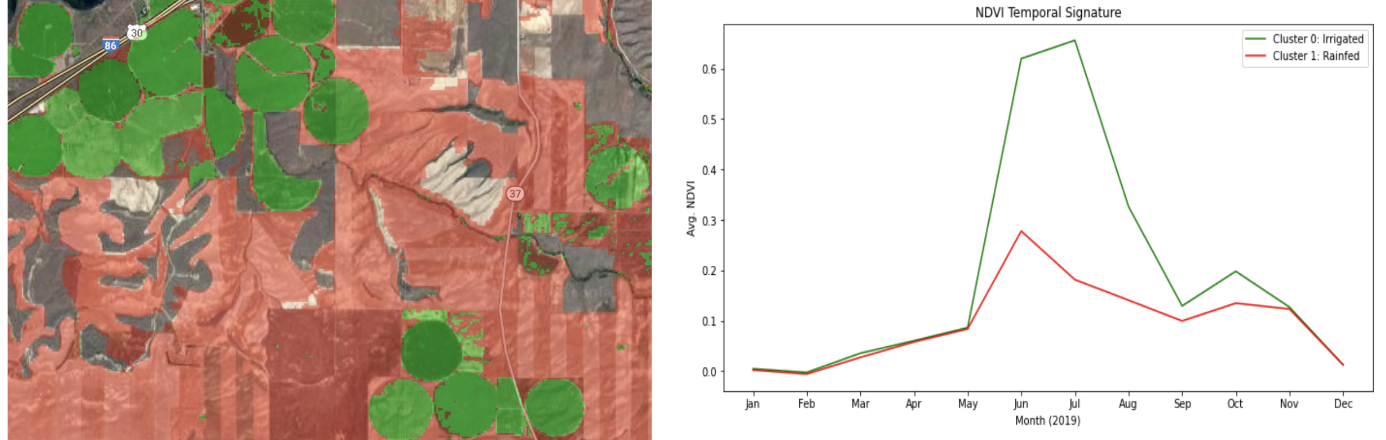}
  \centering
  \caption{(a) 2-cluster K-means output overlaying a Google satellite image, where the green cluster indicates irrigated cropland and the red cluster indicates rainfed cropland. (b) The average NDVI for each month of 2019 by cluster.}
\end{figure}

\subsection{Localized approach}

Performing clustering analysis on the global satellite data at 30m resolution is neither practical due to the size of the data, nor effective due to weak signal-to-noise ratio. To reduce noise, we perform clustering analysis on cropland only, and limit our analysis to bounded regions with edge length of 0.5 degrees. Our approach can be summarized in three steps:

\begin{enumerate}
\item Mask cropland - The GFSAD30 products were used to remove non-cropland areas.
\item Cluster - Clustering analysis was performed on monthly NDVI values on identified cropland areas only. Multiple clustering algorithms were tested, including K-Means, Agglomerative clustering, and Hierarchical Density-Based Spatial Clustering of Applications with Noise (HDBSCAN). Algorithms were evaluated based on the Silhouette Coefficient, the Calinski-Harabasz Index, and the Davies-Bouldin Index. K-Means was chosen for its consistency in performance and reduced compute complexity.
\item Determine cluster labels -  A heuristic approach was used to determine the cluster label for each cluster independent of one another. With this heuristic, we identify crop seasons based on NDVI peaks. NDVI peaks must be at least three months apart to be assigned separate crop seasons (FAO, 1986; Sys et al., 1993). A crop season is labelled as irrigated, as opposed to rainfed, if it satisfies the following criteria:
\begin{enumerate}
    \item The NDVI peak is greater than 0.3 - Kotsuki et al. (2015) assumes that NDVI peak less than 0.4 does not indicate cultivation. The article uses 10-day composite NDVI while monthly NDVI is used in this analysis. Due to the averaging effect, the monthly NDVI peak should be almost always lower than the 10-day NDVI peak. Therefore, a threshold slightly lower than 0.4 was selected.
    \item Average precipitation is less than water need - Average precipitation is calculated using the month before the peak and the month at the peak. There are two reasons why only these two months were taken into consideration. First, the average growing period varies greatly from crop to crop, ranging from three months to eleven months (FAO, 1986). Second, depending on whether the crop is fresh-harvested or dry-harvested, the water need can deviate after the mid-season (FAO, 1986). Without knowledge of the crops planted at the area of interest, only the months before and at the peak are used to calculate the average precipitation. The average monthly water need for crops is usually above 100mm (FAO, 1986). Since there is a large variation of growing periods and water needs not only between crops but also within a crop type, the 100mm threshold was selected by default. However, we reduce this precipitation threshold to 85mm for crop seasons with temperature below 15$^{\circ}$C as there is less water need for cropland growth (FAO, 1986).
\end{enumerate}
\end{enumerate}
Figure 1 shows an example of model predictions over a cropland area in Idaho, USA.

\subsection{Global approach}

An area of interest of any size can be segmented into many small rectangular areas. Each small rectangular area can be modeled in parallel. We ran the model for all of Idaho, USA in parallel on GEE as a proof of concept, and the irrigation predictions can be found on our GEE assets.\footnote{\url{https://code.earthengine.google.com/?asset=users/angela777/test_run_US_ID/edge05_ID}}

\section{Evaluation Methodology and Results}

Given that there are no comprehensive data sources for irrigation at 30m resolution, we conducted two case studies to compare our irrigation predictions against global crowdsourced data from croplands.org and randomly sampled coordinates from GFSAD30’s irrigation products in India and Australia (Gumma et al., 2020). As GFSAD30’s irrigation products are not published datasets and only accessible via a map interface, points must be sampled and assessed manually. 

In our first case study (\#1), we randomly sampled 10 cropland coordinates from GFSAD30’s irrigation products that correspond to labeled crowdsourced coordinates, which acted as our source of truth. At the time of this study, GFSAD30 released 30m resolution irrigation labels exclusively in India and Australia, so our first case study is limited to these geographies. The accuracy of our model is comparable to GFSAD30’s irrigation product, with 70\% and 80\% accuracy, respectively, when compared against the croplands.org crowdsourced labels. 

In our second case study (\#2), we randomly sample 25 geo-diverse cropland coordinates from over 15 countries spanning five continents using the same croplands.org dataset. Our irrigation predictions coincide with crowdsourced labels for 92\% of the coordinates. The results for both case studies are shown in Table 1. More evaluation is required on a larger sample size to increase our confidence in these accuracy levels. 

Lastly, we test the consistency of our model with a custom consistency metric that compares predictions for input regions against an overlapped, spatially-shifted input regions. For each coordinate tested, we compare predictions from an input region against eight translated regions that have been shifted by \nicefrac{1}{3} of the edge length of the input region either vertically or horizontally. Of the points that overlap in the regions, we compare the consistency of irrigation predictions for each 30m pixel. Using 12 randomly sampled global cropland coordinates, our model's predictions are internally consistent for 97\% of the pixels that overlap, indicating that the model is not sensitive to moderate perturbations of the selected input region.

\begin{table}
  \caption{Case study results using croplands.org coordinates}
  \label{results-table}
  \centering
  \begin{tabular}{llll}
    \toprule
    Case Study     & Model     & Accuracy  &  Sample Size \\
    \midrule
    \#1 (India/Aus) & GFSAD30  & 0.80 & N=10     \\
    \#1 (India/Aus)     & Irrigation30 & 0.70 & N=10      \\
    \#2 (Geo-diverse)     & Irrigation30       & 0.92 & N=25  \\
    \bottomrule
  \end{tabular}
\end{table}

\section{Conclusion and Future Work}

With the impact of energy-intensive water usage on climate change, it is essential now more than ever to track global irrigation data at high-resolution. Our model (Irrigation30) demonstrates that unsupervised techniques can be used for predicting the irrigation status of cropland at unprecedented resolution, in any geolocation. With a geographically diverse test set from croplands.org, the Irrigation30 model correctly predicts the irrigation status of 23-of-25 cropland coordinates. While more evaluation is recommended for greater confidence in accuracy levels, we found that our model was insensitive to perturbations of selected input regions, as our custom consistency metric was consistent for 97\% of pixels tested in our third case study. As for future work, we are currently working on self-supervised contrastive learning for irrigation modeling to help improve results. 

\section{Appendix}

\subsection{Acknowledgements}

We thank Prof. Paolo D'Odorico, Lorenzo Rosa, and Prof. Manuela Girotto from the Environmental Sciences department at UC Berkeley. We also extend our thanks to Matthew Brimmer and Adam Sohn from UC Berkeley’s School of Information for supporting and contributing to this project.

\subsection{Countries in Case Study \#2}

Here is a list of countries used in our geo-diverse case study (\#2). 

Lao People’s Democratic Republic, Vietnam, Thailand, Mongolia, United States of America, Indonesia, Kyrgyzstan, Brazil, Tunisia, Malawi, Sri Lanka, Bangladesh, Myanmar, Mali, India, and Ukraine.

\section*{References}

\small

[1] Grafton, R. Q., et al. “The Paradox of Irrigation Efficiency.” {\it Science}, American Association for the Advancement of Science, 24 Aug. 2018, science.sciencemag.org/content/361/6404/748.full.

[2] Tillman, D., et al. Global food demand and the sustainable
intensification of agriculture, Proceedings of the National Academy of Sciences of the United
States of America, November 21, doi: 10.1073/pnas.1116437108.

[3] Deryng, Delphine, et al. “Regional Disparities in the Beneficial Effects of Rising CO2 Concentrations on Crop Water Productivity.” {\it Nature Climate Change}, vol. 6, no. 8, 2016, pp. 786–790., doi:10.1038/nclimate2995.

[4] Mauser, Wolfram, et al. “Global Biomass Production Potentials Exceed Expected Future Demand without the Need for Cropland Expansion.” {\it Nature Communications}, vol. 6, no. 1, 2015, doi:10.1038/ncomms9946.

[5] Meier, Jonas, Florian Zabel, and Wolfram Mauser. "A global approach to estimate irrigated areas–a comparison between different data and statistics." {\it Hydrology and Earth System Sciences} 22.2 (2018): 1119

[6] Yan, L., and D. P. Roy. "Conterminous United States crop field size quantification from multi-temporal Landsat data." {\it Remote Sensing of Environment} 172 (2016): 67-86.

[7] Fritz, Steffen, et al. "Mapping global cropland and field size." {\it Global change biology} 21.5 (2015): 1980-1992.

[8] Xiong, J., Thenkabail, P. S., Tilton, J.C., Gumma, M. K., Teluguntla, P., Congalton, R. G., Yadav, K., Dungan, J., Oliphant, A. J., Poehnelt, J., Smith, C., Massey, R. (2017). NASA Making Earth System Data Records for Use in Research Environments (MEaSUREs) Global Food Security-support Analysis Data (GFSAD) @ 30-m Africa: Cropland Extent Product (GFSAD30AFCE). NASA EOSDIS Land Processes DAAC. Retrieved from https://doi.org/10.5067/MEaSUREs/GFSAD/GFSAD30AFCE.001

[9] Xiong, J.; Thenkabail, P.S.; Tilton, J.C.; Gumma, M.K.; Teluguntla, P.; Oliphant, A.; Congalton, R.G.; Yadav, K.; Gorelick, N. Nominal 30-m Cropland Extent Map of Continental Africa by Integrating Pixel-Based and Object-Based Algorithms Using Sentinel-2 and Landsat-8 Data on Google Earth Engine. Remote Sens. 2017, 9, 1065. doi:10.3390/rs9101065 ; http://www.mdpi.com/2072-4292/9/10/1065 Also, the cover story: http://www.mdpi.com/2072-4292/9/10

[10] Teluguntla, P., Thenkabail, P.S., Xiong, J., Gumma, M.K., Congalton, R. G., Oliphant, A. J., Sankey, T., Poehnelt, J., Yadav, K., Massey, R., Phalke, A., Smith, C. (2017). NASA Making Earth System Data Records for Use in Research Environments (MEaSUREs) Global Food Security-support Analysis Data (GFSAD) @ 30-m for Australia, New Zealand, China, and Mongolia: Cropland Extent Product (GFSAD30AUNZCNMOCE). NASA EOSDIS Land Processes DAAC. Retrieved from https://doi.org/10.5067/MEaSUREs/GFSAD/GFSAD30AUNZCNMOCE.001

[11] Oliphant, A. J., Thenkabail, P. S., Teluguntla, P., Xiong, J. Congalton, R. G., Yadav, K., Massey, R., Gumma, M.K., Smith, C. 2017. NASA Making Earth System Data Records for Use in Research Environments (MEaSUREs) Global Food Security-support Analysis Data (GFSAD) @ 30-m for Southeast \& Northeast Asia: Cropland Extent Product (GFSAD30SEACE). NASA EOSDIS Land Processes DAAC. Retrieved from https://doi.org/10.5067/MEaSUREs/GFSAD/GFSAD30SEACE.001

[12] Gumma, M.K., Thenkabail, P.S., Teluguntla, P., Oliphant, A.J., Xiong, J., Congalton, R. G., Yadav, K., Phalke, A., Smith, C. (2017). NASA Making Earth System Data Records for Use in Research Environments (MEaSUREs) Global Food Security-support Analysis Data (GFSAD) @ 30-m for South Asia, Afghanistan and Iran: Cropland Extent Product (GFSAD30SAAFGIRCE). NASA EOSDIS Land Processes DAAC. Retrieved fromhttps://doi.org/10.5067/MEaSUREs/GFSAD/GFSAD30SAAFGIRCE.001

[13] Phalke, A., Ozdogan, M., Thenkabail, P. S., Congalton, R. G., Yadav, K., Massey, R., Teluguntla, P., Poehnelt, J., Smith, C. (2017). NASA Making Earth System Data Records for Use in Research Environments (MEaSUREs) Global Food Security-support Analysis Data (GFSAD) @ 30-m for Europe, Middle-east, Russia and Central Asia: Cropland Extent Product (GFSAD30EUCEARUMECE). NASA EOSDIS Land Processes DAAC. Retrieved fromhttps://doi.org/10.5067/MEaSUREs/GFSAD/GFSAD30EUCEARUMECE.001

[14] Massey, R., Sankey, T.T., Yadav, K., Congalton, R.G., Tilton, J.C., Thenkabail, P.S. (2017). NASA Making Earth System Data Records for Use in Research Environments (MEaSUREs) Global Food Security-support Analysis Data (GFSAD) @ 30m for North America: Cropland Extent Product (GFSAD30NACE). NASA EOSDIS Land Processes DAAC. https://doi.org/10.5067/MEaSUREs/GFSAD/GFSAD30NACE.001

[15] Zhong, Y., Giri, C., Thenkabail, P.S., Teluguntla, P., Congalton, R. G., Yadav, K., Oliphant, A. J., Xiong, J., Poehnelt, J., and Smith, C. 2017. NASA Making Earth System Data Records for Use in Research Environments (MEaSUREs) Global Food Security-support Analysis Data (GFSAD) @ 30-m for South America: Cropland Extent Product (GFSAD30SACE). NASA EOSDIS Land Processes DAAC. Retrieved fromhttps://doi.org/10.5067/MEaSUREs/GFSAD/GFSAD30SACE.001

[16] Congalton, R. G., Yadav, K., McDonnell, K., Poehnelt, J., Stevens, B., Gumma, M. K., Teluguntla, P., Thenkabail, P.S. (2017). NASA Making Earth System Data Records for Use in Research Environments (MEaSUREs) Global Food Security-support Analysis Data (GFSAD) @ 30-m: Cropland Extent Validation (GFSAD30VAL). NASA EOSDIS Land Processes DAAC. Retrieved from https://doi.org/10.5067/MEaSUREs/GFSAD/GFSAD30VAL.001

[17] P. Thenkabail, P. Teluguntla, J Xiong, A. Oliphant, R. Massey (2016). NASA MEaSUREs Global Food Security Support Analysis Data (GFSAD) Crop Mask 2010 Global 1 km V001. NASA EOSDIS Land Processes DAAC. Retrieved from https://doi.org/10.5067/MEaSUREs/GFSAD/GFSAD1KCM.001

[18] Xie, Yanhua, et al. "Mapping irrigated cropland extent across the conterminous United States at 30 m resolution using a semi-automatic training approach on Google Earth Engine." {\it ISPRS Journal of Photogrammetry and Remote Sensing} 155 (2019): 136-149.

[19] Oliphant, Adam J., et al. "Mapping cropland extent of Southeast and Northeast Asia using multi-year time-series Landsat 30-m data using a random forest classifier on the Google Earth Engine Cloud." {\it International Journal of Applied Earth Observation and Geoinformation} 81 (2019): 110-124.

[20] Gumma, Murali Krishna, et al. "Agricultural cropland extent and areas of South Asia derived using Landsat satellite 30-m time-series big-data using random forest machine learning algorithms on the Google Earth Engine cloud." {\it GIScience \& Remote Sensing} 57.3 (2020): 302-322.

[21] Phalke, Aparna, et al. "Mapping croplands of Europe, Middle East, Russia, and Central Asia using Landsat 30-m data, machine learning algorithms and Google Earth Engine." {\it ISPRS Journal of Photogrammetry and Remote Sensing} 167 (2020): 104-122.

[22] Abatzoglou, J.T., S.Z. Dobrowski, S.A. Parks, K.C. Hegewisch, 2018, Terraclimate, a high-resolution global dataset of monthly climate and climatic water balance from 1958-2015, Scientific Data 5:170191, doi:10.1038/sdata.2017.191

[23] FAO – Food and Agriculture Organization of the United Nations: "Irrigation water management: irrigation water needs." {\it Training manual} 3, http://www.fao.org/3/S2022E/s2022e00.htm (last access: 4 October 2020), 1986. 

[24] Sys, C. O., van Ranst, E., Debaveye, J., and Beernaert, F.: Land Evaluation: Part III Crop Requirements, Agricultural Publications, General Administration for Development Cooperation, Brussels, 1993. 

[25] Kotsuki, S., and K. Tanaka. "SACRA-a method for the estimation of global high-resolution crop calendars from a satellite-sensed NDVI." (2015).


\end{document}